\documentclass[letterpaper, 10 pt, conference]{ieeeconf}  %

\IEEEoverridecommandlockouts                              %

\usepackage{amsmath} %
\usepackage{amssymb}  %

\usepackage{bm}  %
\usepackage{graphicx}  %
\usepackage{subcaption} %
\usepackage{siunitx}

\usepackage{algpseudocode} %
\usepackage{algorithm} %
\usepackage[T1]{fontenc} %
\usepackage{wrapfig} %

\usepackage[
        backend=biber,
        style=numeric,
        sorting=none,
        doi=false,
        isbn=false,
        url=false]{biblatex}

\usepackage{listings}

\title{\LARGE \bf
        Inverse Kinematics with Forward Dynamics Solvers for Sampled Motion Tracking}

\author{Stefan Scherzinger$^{1}$, Arne Roennau$^{1}$ and R\"udiger Dillmann$^{2}$%
\thanks{$^{1}$Stefan Scherzinger and Arne Roennau are with FZI Research Center for Information Technology, Haid-und-Neu-Str. 10-14, 76131 Karlsruhe, Germany
{\tt\small stefan.scherzinger@fzi.de} {\tt\small arne.roennau@fzi.de}}%
\thanks{$^{2}$R\"udiger Dillmann is with IAR Institute for Anthropomatics and Robotics, HIS Humanoids and Intelligence Systems Lab, KIT Karlsruhe Institute of Technology,
        Adenauerring 2, 76131 Karlsruhe, Germany
{\tt\small ruediger.dillmann@kit.edu}}%
}

\begin{document}

\onecolumn
\noindent
\textsuperscript{\textcopyright}2019 IEEE.  Personal use of this material is permitted.  Permission from IEEE must be obtained for all other uses, in any current or future media, including reprinting/republishing this material for advertising or promotional purposes, creating new collective works, for resale or redistribution to servers or lists, or reuse of any copyrighted component of this work in other works.

\vspace{0.5cm}
\noindent
Please cite this paper as:
 
\lstset{%
  basicstyle=\fontsize{9}{10}\selectfont\ttfamily
}%
\begin{lstlisting}
@INPROCEEDINGS{Scherzinger2019Inverse,
author={S. {Scherzinger} and A. {Roennau} and R. {Dillmann}},
booktitle={19th International Conference on Advanced Robotics (ICAR)},
title={Inverse Kinematics with Forward Dynamics Solvers for Sampled Motion Tracking},
year={2019},
volume={},
number={},
pages={681-687},
keywords={},
doi={10.1109/ICAR46387.2019.8981554},
ISSN={},
month={Dec},}
\end{lstlisting}

\twocolumn
\newpage

\maketitle
\thispagestyle{empty}
\pagestyle{empty}

\begin{abstract}
Tracking Cartesian motion with end~effectors is a fundamental task in robot control.
For motion that is not known in advance, the solvers must find
fast solutions to the inverse kinematics (IK) problem for discretely sampled target poses.
On joint control level, however, the robot's actuators operate in a continuous domain,
requiring smooth transitions between individual states.
In this work, we present a boost to the well-known Jacobian transpose method
to address this goal, using the mass matrix of a virtually conditioned
twin of the manipulator.
Results on the UR10 show superior convergence and quality of our dynamics-based solver against the plain Jacobian method.
Our algorithm is straightforward to implement as a controller, using common robotics libraries.
\end{abstract}

\section{INTRODUCTION}  %
In robotics, solutions to the Inverse Kinematics (IK) problem are fundamental for manipulator control.
Articulated mechanisms are powered in their joints, and hence
require control algorithms to perform continuous mappings from task space
to joint space.
Being such a core element in robot control, IK has received significant
attention in the last decades and has lead to a wide range of different
methods. See e.g. \cite{Aristidou2018} for an overview.
Among the fastest approaches are closed form solutions, which are analytically
derived solutions, specifically tailored for each robot kinematics.
\textit{IKFast}\footnotemark[3]
 with an example application in \cite{Diankov2011}.
However, closed form solutions are sensitive to impossible-to-reach targets.
Numerical approaches provide trade-offs between accuracy and
stability, usually leveraging the manipulator Jacobian: E.g. its transpose
\cite{Wolovich1984}, \cite{Balestrino1984}, \cite{Pechev2008}, its inverse
with singularity stabilizing measures, such as Damped Least Squares (DLS)
\cite{Nakamura1986}, \cite{Wampler1988}, \cite{Beeson2015}
or Selectively Damped Least Squares (SDLS) \cite{Buss2005}.
More recent improvements have shown excellent success rates for general IK solving \cite{Beeson2015}.
While suitable for searching vast regions of the solution space  in the context of
\textit{motion planning}, finding individual joint configurations is not always sufficient:

Target motion is often sampled into discrete target poses, such as in virtual servoing or end~effector teleoperation.
Solving the IK problem for each target individually leads to valid, but decoupled solutions in joint space with no
guaranteed feasibility of execution.
On the other hand, for low-frequency sampled targets, individual solutions are too sparse
to serve as direct joint control commands and cause jumps in the robot's actuators.
A remedy is interpolating in task space, leading again to the problem of decoupled solutions in joint space.%

In this paper we describe an IK approach that inherently achieves both goals for real-time target following.
Our approach enhances the well-known Jacobian transpose method with a simple
but effective component -- a conditioned, virtual mass matrix -- leading to intuitive solutions of the IK solver and to smooth
intermediate states between sparse targets.
To support the vastly growing Robot Operating System (ROS)\footnotemark[4] \cite{Quigley2009}, we
accompany the paper with a \textit{ROS-control} \cite{Chitta2017} controller implementation as power-on-and-go
solution for the robotics community\footnotemark[5].

\section{Problem statement and related work}
\footnotetext[3]{http://openrave.org/docs/0.8.2/openravepy/ikfast/}
\footnotetext[4]{www.ros.org}
\footnotetext[5]{https://github.com/fzi-forschungszentrum-informatik/cartesian\_controllers}
The forward mapping of the joint positions vector $\bm{q}$ to Cartesian space is given by
\begin{equation}
        \label{eq:forward_kinematics}
                \bm{x} = g \left( \bm{q} \right)
\end{equation}
such that $\bm{q} = g^{-1}(\bm{x})$ would represent a close form solution to the inverse problem,
which is to be approximated numerically.
The velocity vector $\dot{\bm{q}}$ maps with
\begin{equation}
        \label{eq:manipulator_jacobian}
                \dot{\bm{x}} = \bm{J} \left( \bm{q} \right) \dot{\bm{q}}
\end{equation}
to end~effector velocity $\dot{\bm{x}}$, using the manipulator Jacobian $\bm{J}$.
Furthermore, the common relation
\begin{equation}
        \label{eq:joint_torques}
                \bm{\tau} = \bm{J}^{T} \left( \bm{q} \right) \bm{f}
\end{equation}
holds for a static end~effector force vector $\bm{f}$ and torques $\bm{\tau}$ in the joints of the robot manipulator.
In further notations we will drop the joint configuration dependency of $\bm{J}(\bm{q})$ for the sake of brevity. 
Using $\bm{J}^T$ to solve (\ref{eq:forward_kinematics}) for $\bm{q}$ has been proposed independently by Wolovich et al~\cite{Wolovich1984} and Balestrino et al~\cite{Balestrino1984}.

In \cite{Balestrino1984} the authors describe
a dynamical systems according to
\begin{equation}
        \label{eq:balestrino_dynamic_system}
                \dot{\bm{q}} = \bm{J}^T  \bm{Q} (\bm{x}^d - g(\bm{q})) + \bm{w}
\end{equation}

for coordinate transformations from task- to joint space,
with $\bm{Q}$ denoting an arbitrary positive definite matrix and $\bm{w} \left( \bm{Q}, \bm{J}, \bm{\epsilon} \right)$ representing an adaptation term, whose exact expression we do not repeat here.
The authors show asymptotic stability around the Cartesian desired pose $\bm{x}^d$ with $\bm{\epsilon} = \bm{x}^d - g(\bm{q})$,
and validate their approach on a three DOF mechanism.
Although proving effective in converging, (\ref{eq:balestrino_dynamic_system}) does not offer an intuitive interpretation.
Furthermore, the authors state that velocity tracking is only matched in the
mean, requiring additional filters to deliver real angular velocities for robot control.

In \cite{Pechev2008} Pechev uses a similar technique with Feedback Inverse Kinematics (FIK) from a control perspective,
deriving
\begin{equation}
        \label{eq:pechev_dynamic_system}
                \dot{\bm{q}} = \bm{J}^T \bm{Q} \left( \bm{J} \bm{J}^T \bm{Q} + \bm{I} \right)^{-1} \dot{\bm{x}}^d
\end{equation}
as system dynamics.
In this case $\bm{Q}$ denotes a dynamic, non-diagonal matrix that involves an off-line state space optimization
subject to manipulator dynamics and task requirements.
Solving (\ref{eq:pechev_dynamic_system}) in a feedback loop as a filter avoids matrix inversions, and performs well
in singularity experiments in comparison to the DLS method from \cite{Nakamura1986}, albeit
only presented on a three DOF planar manipulator.
The Jacobian transpose is also commonly used in Cartesian control, see e.g. \cite{Craig2009}, using
the real plants dynamics in these cases.
Here, however, we separate the computation of the IK problem from the
underlying manipulator, which we see as a black box to be controlled in an open loop fashion
with desired configurations $\bm{q}^d(t)$ exclusively.

In~\cite{Wolovich1984} the authors derive a solution to the IK problem by proposing the usage of
$\bm{J}^{T}$ in form of a simpler dynamical system
\begin{equation}
        \label{eq:wolovich_dynamic_system}
                \ddot{\bm{q}} = \bm{K} \bm{J}^T  \left( \bm{x}^d - g(\bm{q}) \right)
\end{equation}
They provide a Lyapunov stability analysis and show that the system is asymptotically stable in the error dynamics,
using an arbitrary positive definite matrix $\bm{K}$.
However, they leave a possible conditioning of $\bm{K}$ aside and remain with the general proof.
In the remainder of the paper we will refer to this approach as the \textit{Jacobian transpose method}.

In this work, we investigate the benefits of choosing $\bm{K}$ from
(\ref{eq:wolovich_dynamic_system}) not to be constant, but instead to be the dynamically changing
inertia matrix of a virtually conditioned twin of the robot manipulator.
Like the work in \cite{Pechev2008} we show the importance of matrices with off-diagonal elements,
but introduce a more intuitive interpretation, using manipulator dynamics.
A primary contribution of this paper is to propose \textit{forward dynamics
solvers} in the context of $\bm{J}^T$-based IK solving.
This general idea bases on previous findings \cite{Scherzinger2017}.
Here, we present an enhanced algorithm with improved convergence with
physically plausible intermediate solutions for motion control.

\section{FORWARD DYNAMICS IK SOLVER}       %

\subsection{Simulation of Robot Motion}
Approximating the manipulator as a set of articulated, rigid bodies, the
governing equations of motion obtain the following form in matrix notation
\begin{equation}
        \label{eq:rigid_body_system}
        \bm{\tau} = \bm{H}({\bm{q}}) \ddot{\bm{q}} + \bm{C}( \bm{q},\dot{\bm{q}} ) + \bm{G}(\bm{q})
\end{equation}
in which $\bm{H}$ denotes the mechanism's positive definite inertia matrix and
$\bm{C}$ and $\bm{G}$ represent vectors with separated Coriolis an
centrifugal terms and gravitational components respectively.
Rearranging for $\ddot{\bm{q}}$ and using (\ref{eq:joint_torques}) yields the manipulator's acceleration due to external forces $\bm{f}$ acting on its end~effector.
\begin{equation}
        \label{eq:forward_dynamics}
        \ddot{\bm{q}} = \bm{H}^{-1} \bm{J}^{T} \bm{f} - \bm{H}^{-1} \bm{C}( \bm{q},\dot{\bm{q}} ) - \bm{H}^{-1} \bm{G}(\bm{q})
\end{equation}
Again, we dropped the dependency of $\bm{q}$ in the notation for brevity.
Solving (\ref{eq:forward_dynamics}) for $\bm{q}$ is considered with \textit{forward
dynamics}, i.e. simulating the mechanism's motion through time if
loads $\bm{f}$ are applied. This is commonly achieved with numerical
integration methods and is a vital part in physics engines.
The right-hand side of (\ref{eq:forward_dynamics}) is a superposition of acceleration terms in joint space.
While taking all terms into consideration is crucial for simulating physically correct motion,
our approach for solving IK problems has a different focus and motivates two simplifications:
If we drop acceleration due to gravity ($\bm{G} = \bm{0}$) and consider instantaneous motion only for each cycle (i.e. $\dot{\bm{q}} =
\bm{0}$), which equals setting $\bm{H}^{-1} \bm{C}( \bm{q},\dot{\bm{q}} ) =
\bm{0}$, our system still accelerates in the direction pointed to by $\bm{f}$,
but that acceleration is not biased by velocity-dependent non-linearities, nor
the effect of gravity.
To be concise with this simplification, we do not accumulate velocity during time integration in our algorithm in section~\ref{sec:implementation}.
Implications of this approach are discussed in section~\ref{sec:discussion}.
Expelling both acceleration terms from (\ref{eq:forward_dynamics}), we obtain the simplified system
\begin{equation}
        \label{eq:simplified_system}
        \ddot{\bm{q}} = \bm{H}^{-1} \bm{J}^{T} \bm{f} ~,
\end{equation}
which relates $\bm{f}$ to instantaneous joint acceleration $\ddot{\bm{q}}$.
Turning the search for IK solutions into a control problem,
(\ref{eq:simplified_system}) is the control plant we chose at the core of our
solver.

We use the distance error $\bm{\epsilon} = \bm{x}^d - g(\bm{q})$ as the error between target and current end~effector pose.
Its entries are
\begin{equation}
        \label{eq:pose_error}
        \bm{\epsilon} = \left[ \epsilon_x, \epsilon_y, \epsilon_z, \epsilon_{rx}, \epsilon_{ry}, \epsilon_{rz} \right]^T
\end{equation}
with the translational error components and rotational error components of the according Rodrigues vector.

\begin{figure}
        \centering
        \includegraphics[width=.48\textwidth]{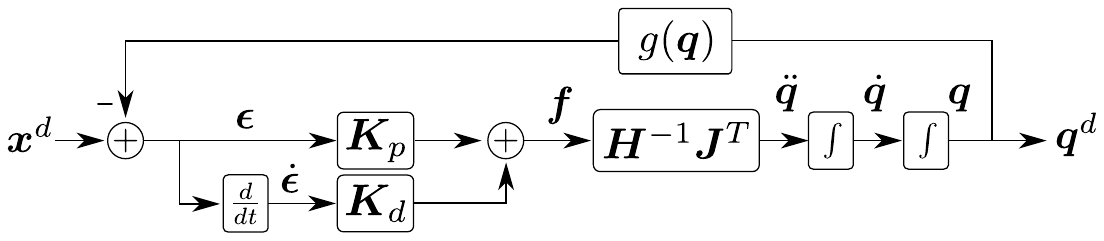}
        \caption{Closed loop control scheme of our IK solver. The control plant uses simplified forward dynamics computation of a virtually-conditioned twin of the real robot.}
        \label{fig:fd_ik_solver}
\end{figure}

Relating this distance error
vector to a virtual Cartesian end~effector input
\begin{equation}
        \label{eq:force_control}
        \bm{f} =  \bm{K}_p \bm{\epsilon} + \bm{K}_d \dot{\bm{\epsilon}}
\end{equation}
with the positive definite, diagonal gain matrices $\bm{K}_p$ and $\bm{K}_d$,
we obtain
\begin{equation}
        \label{eq:fd_ik_solver}
        \ddot{\bm{q}} = \bm{H}^{-1} \bm{J}^{T} \left( \bm{K}_p \bm{\epsilon} + \bm{K}_d \dot{\bm{\epsilon}} \right)
\end{equation}
as our forward dynamics motivated controller.
Fig. \ref{fig:fd_ik_solver} shows the closed loop control scheme for solving IK.

Note the similarity to (\ref{eq:wolovich_dynamic_system}): Setting $\bm{K}_p = \bm{I}$ and $\bm{K}_d~=~\bm{0}$ yields
\begin{equation}
        \label{eq:comparison_wolovich}
        \ddot{\bm{q}} = \bm{H}^{-1} \bm{J}^T  \left( \bm{x}^d - g(\bm{q}) \right)~~.
\end{equation}
However, $\bm{H}^{-1}$ generates physically plausible joint accelerations
$\ddot{\bm{q}}$ that really reflect the mechanism reaction to the Cartesian error vector for instantaneous motion.
This means that meeting kinematic constraints, the mechanism accelerates in Cartesian space with $\ddot{\bm{x}}$ in
the direction as pointed to by $\bm{f}$. In the experiments section we show
that this indeed causes faster and more goal-directed convergence.
\subsection{Homogenization Methods}
A goal of this paper is to provide insight on the conditioning of the mechanism
such that $\bm{H}(\bm{q})$ possesses beneficial behavior throughout the joint
space. Note that unlike the kinematics of the system, which we can not change,
we are free to give the simulated manipulator any dynamic behavior we wish.
In this context we propose to think of the solver concept from
Fig.~\ref{fig:fd_ik_solver} as using a virtually conditioned twin of the
real mechanism for which we wish to solve IK.
The basic idea behind this approach has been presented in our previous work~\cite{Scherzinger2017}.
In this paper, we present more detailed aspects behind this concept and provide
an empirical proof of this earlier proposition.

The time derivative of (\ref{eq:manipulator_jacobian}) gives
\begin{equation}
        \label{eq:jacobian_diff}
                \ddot{\bm{x}} = \dot{\bm{J}} \dot{\bm{q}} + \bm{J} \ddot{\bm{q}}~~.
\end{equation}
We again consider instantaneous accelerations while the mechanism is still at
rest and set $\dot{\bm{J}} \dot{\bm{q}} = \bm{0}$.
Together with (\ref{eq:simplified_system}) we obtain
\begin{equation}
        \label{eq:cartesian_acceleration}
        \ddot{\bm{x}} = \bm{J} \bm{H}^{-1} \bm{J}^T \bm{f}~~,
\end{equation}
which describes the Cartesian instantaneous acceleration due to the controlled distance error $\bm{f}$.
With setting
$\bm{M}^{-1} = \bm{J} \bm{H}^{-1} \bm{J}^T$
we obtain the more concise notation
\begin{equation}
        \label{eq:cartesian_acceleration_short}
        \ddot{\bm{x}} = \bm{M}^{-1} \bm{f}~~.
\end{equation}
\begin{wrapfigure}{r}{0.20\textwidth}
        \includegraphics[width=0.20\textwidth]{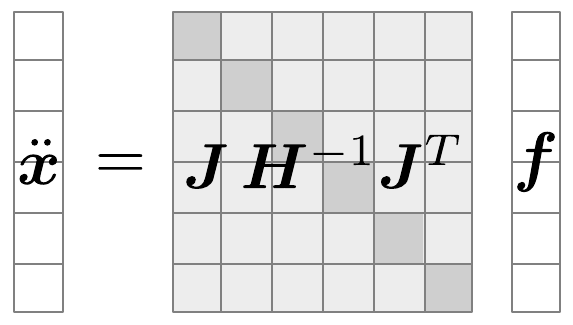}
\end{wrapfigure}
Our intention now is to achieve $\bm{M}^{-1} (\bm{q})$ to be an almost
constant, diagonal matrix for all possible joint configurations $\bm{q}$, as illustrated on the right.
So that the gain matrices $\bm{K}_p$
and $\bm{K}_d$ generate uniform system accelerations $\ddot{\bm{x}}$, and our control loop from
Fig.~\ref{fig:fd_ik_solver} has equal convergence throughout the Cartesian workspace.

\begin{figure}
        \centering
        \includegraphics[width=.35\textwidth]{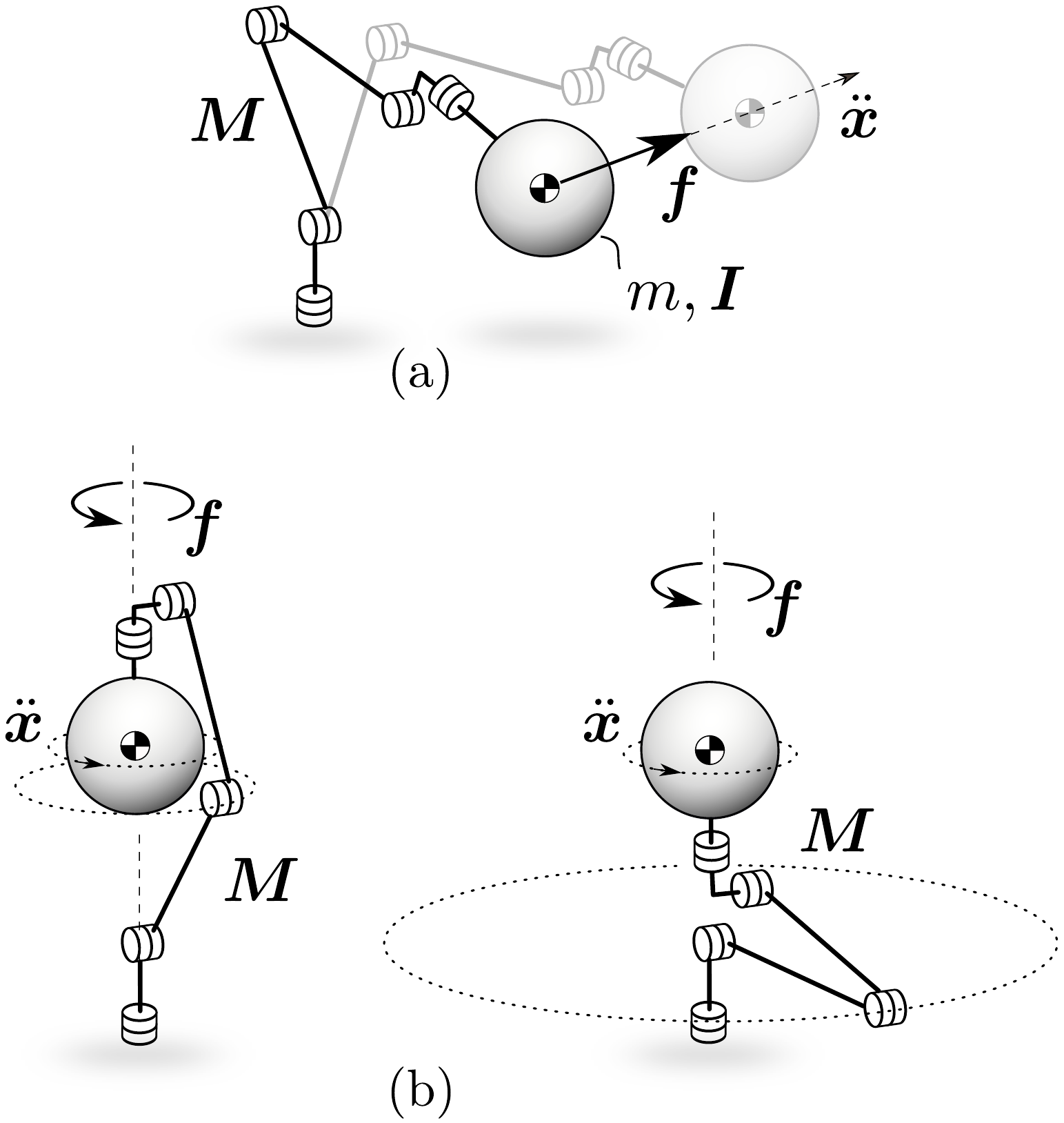}
        \caption{Dynamics-conditioned twin of an exemplary mechanism. The
        end~effector comprises all of the mechanism's mass $m$ and rotational inertia $\bm{I}$, which
        is illustrated with the oversized sphere. Links and joints are
        considered massless, such that their configuration has a vanishing influence on the overall dynamics.}
        \label{fig:virtual_mechanism}
\end{figure}

To achieve this behavior, we follow a mechanically-motivated assumption: The error-correcting $\bm{f}$ acts directly on
the mechanism's end~effector.  If this end~effector comprises all of the
mechanism's mass and inertia, as illustrated in
Fig.~\ref{fig:virtual_mechanism}a, then joint configurations will have less
effects on $\bm{M}$. The overall center of mass remains unchanged.
This effect is depicted in Fig.~\ref{fig:virtual_mechanism}b where $\bm{f}$
experiences the same rotational inertia $\bm{M}$, although having shifted most
of the mechanism's structure away from the rotatory axis in one case.
In the experiments section we sample a massive number of joint configurations and show empirically that this approach indeed
achieves a good homogenization of $\bm{M}^{-1} = \bm{J} \bm{H}^{-1} \bm{J}^T$.

\section{IMPLEMENTATION}       %
\label{sec:implementation}
We implemented the closed loop scheme from
Fig.~\ref{fig:fd_ik_solver} as a \textit{ROS controller} in C++. The pseudo code for the IK-solving part is listed in Algorithm~\ref{alg:solver_algorithm}.
\begin{algorithm}[H]
        \caption{IK with forward dynamics}
        \label{alg:solver_algorithm}
        \begin{algorithmic}[1]
        \Procedure{Solver}{$\bm{x}^d, \bm{q}_0, \Delta t, N$}
                \State $\bm{\epsilon}_0 = \bm{0}$
                \For{$i = 1 \text{ \bf{to} } N$}
                        \State $\bm{\epsilon}_i = \bm{x}^d - g(\bm{q}_{i-1})$
                        \State $\dot{\bm{\epsilon}}_i = (\bm{\epsilon}_i - \bm{\epsilon}_{i-1}) / \Delta t$
                        \State $\bm{f}_i = \bm{K}_p \bm{\epsilon}_i + \bm{K}_d \dot{\bm{\epsilon}}_i$
                        \State $\ddot{\bm{q}}_i = \bm{H}^{-1}(\bm{q}_{i-1}) \bm{J}^T (\bm{q}_{i-1}) \bm{f}_i$
                        \State $\dot{\bm{q}}_i = 0.5~\ddot{\bm{q}}_i \Delta t$ \Comment{$\ddot{\bm{q}}_{i-1} \equiv \bm{0}$}
                        \State $\bm{q}_i = \bm{q}_{i-1} + 0.5~ \dot{\bm{q}}_i \Delta t$ \Comment{$\dot{\bm{q}}_{i-1} \equiv \bm{0}$}
                \EndFor
                \State $\bm{q}^d = \bm{q}_N$
                \State \textbf{return} $\bm{q}^d$
        \EndProcedure
        \end{algorithmic}
\end{algorithm}
Note that the computation of $\bm{H}^{-1}$ and $\bm{J}^T$ is not part of the algorithm. For this
task we used the Kinematics and Dynamics Library (KDL)\footnote{
http://wiki.ros.org/orocos\_kinematics\_dynamics}, which is common place in ROS.
Note that the number of steps $N$ can be chosen freely to give the solver
(virtual) time for any given $\bm{x}^d$. In combination with the
gain matrices $\bm{K}_p$ and $\bm{K}_d$ this is a partially redundant means to
tweak the solver to range from a one-shot IK-solver to an interpolating controller to smooth noisy targets $\bm{x}^d$.
We discuss this behavior in the experiment on interpolation performance for low frequency sampled targets.

\section{EXPERIMENTS AND RESULTS}       %
\begin{wrapfigure}{r}{0.17\textwidth}
        \includegraphics[width=0.17\textwidth]{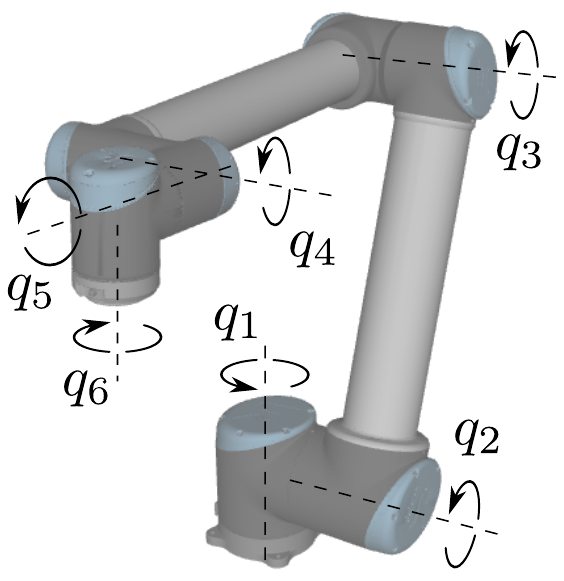}
        \caption{UR10}
        \label{fig:ur10}
\end{wrapfigure}

In our experiments, we chose the Universal Robot UR10 as manipulator (Fig.~\ref{fig:ur10}), which
is ubiquitous both in industry and academia, and represents a well-known platform for many possible users.
We want to emphasize, however, that our proposed IK solver has been used successfully on various manipulators.
Since a central theme of this work is to use a dynamics motivated approach for IK
solving instead of a pure kinematics one, we compare our method to the
Jacobian transpose method where appropriate.

\subsection{Controller Homogenization}
In this experiment we evaluated our homogenization method and compared
it both to the Jacobian transpose method
from (\ref{eq:wolovich_dynamic_system}) and a
reference dynamics model.
For all mechanism, we used the robot's kinematics as provided by the ROS package\footnote{https://github.com/ros-industrial/universal\_robot}.
\begin{figure}
        \centering
        \begin{subfigure}[b]{0.4\textwidth}
                \includegraphics[width=1.0\textwidth]{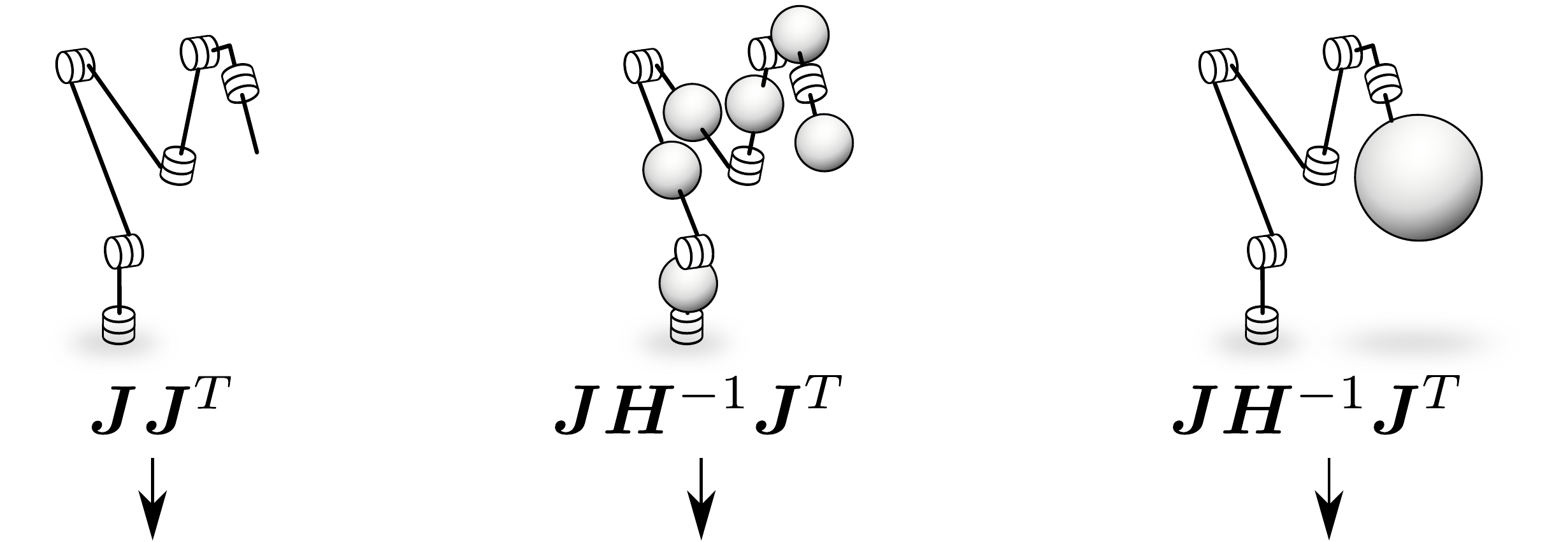}
                \caption{}
                \label{fig:homogenization_models} 
        \end{subfigure}

        \begin{subfigure}[b]{0.490\textwidth}
                \includegraphics[width=1.0\textwidth]{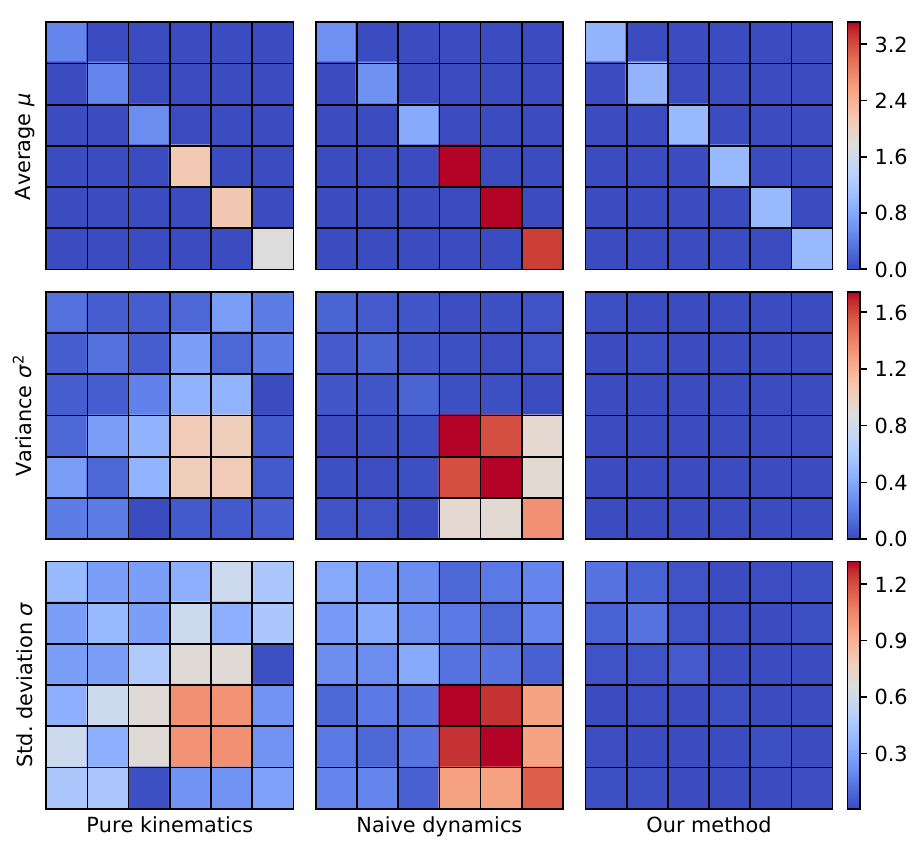}
                \caption{}
                \label{fig:homogenization_statistics}
        \end{subfigure}
        \caption{%
                \textbf{Experiment A}. Analysis of three mapping approaches from $\bm{f} \rightarrow \ddot{\bm{x}}$ for the Universal
                Robots UR10.
                To obtain the plots, we sampled $100.000$ random configurations of $\bm{q}$
                uniformly with $q_i \in \left[-\pi, \pi \right]$.
                For each sample, we computed the mapping matrix as indicated in (a).
                The sub figures (b) show mean, variance and standard
                deviation (less is better) for the individual matrix entries for the entirety of samples.
        }
        \label{fig:homogenization}
\end{figure}
For the Jacobian transpose method
, we set the gain matrix $\bm{K}$ to identity, which is a reasonable
choice without further knowledge of the system.

Fig.~\ref{fig:homogenization_models} illustrates the qualitative distribution of mass and inertia for the experiment.
We choose the following values for our proposed model (Fig.~\ref{fig:homogenization_models} right):
\begin{equation}
        \label{eq:mass_and_inertia}
        \begin{aligned}
                m &= \SI{1}{kg} \\
                \bm{I} &= \begin{bmatrix}
                        1 & 0 & 0 \\
                        0 & 1 & 0 \\
                        0 & 0 & 1
                \end{bmatrix}~ \SI{}{kgm^2}
        \end{aligned}
\end{equation}
which constitute the last link of the manipulator.
For numerical stability, we set all other links to $m_0 = 10^{-3} m$ and $\bm{I}_0 = 10^{-6} \bm{I}$ respectively (instead of to zero).
The reference model (Fig.~\ref{fig:homogenization_models} middle) had \num{1/6} of $m$ and $\bm{I}$ equally attached to its links.

Fig.~\ref{fig:homogenization_statistics} shows the results of the analysis.
Note that all mean matrices are in fact diagonal, as was expected for arbitrary joint space sampling.
However, the variances indicate a partially strong configuration dependency for
Jacobian transpose method (pure kinematic solver) and the equally conditioned mechanism (naive dynamics), indicating
suboptimal solver convergence if they were applied to solve IK.
Note that using dynamics does not necessarily improve homogenization (middle
column). Only with the end~effector approach (right column) do we effectively obtain the intended behavior.

We propose the values from (\ref{eq:mass_and_inertia}) to be considered as good default values for a broad usage. We also take these values for the following experiments of this paper.

\subsection{Solver Convergence}
\begin{figure*}
        \centering
        \begin{subfigure}[b]{0.92\textwidth}
                \includegraphics[width=1.0\textwidth]{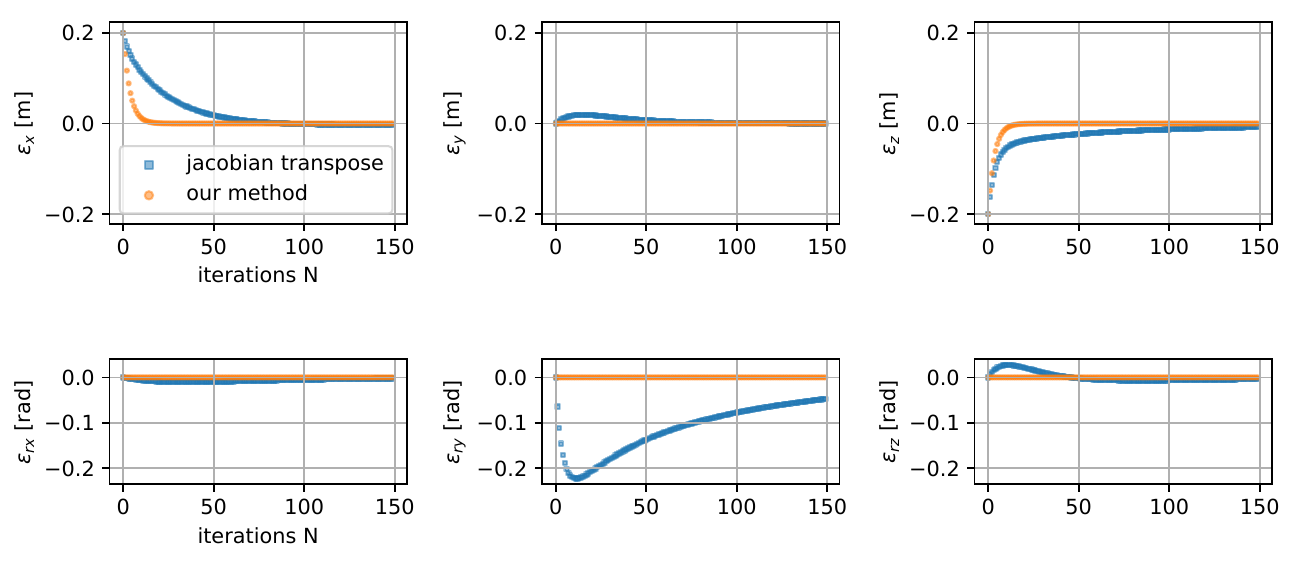}
                \label{fig:solver_convergence_plots} 
        \end{subfigure}%
        \caption{
                \textbf{Experiment B}. Analysis of the solver convergence for the Jacobian method: $\ddot{\bm{q}} = \alpha \bm{I}_6 \bm{J}^T
                \bm{K}_p (\bm{x}^d - g(\bm{q}))$,
                and our method: $\ddot{\bm{q}} = \bm{H}^{-1} \bm{J}^T \bm{K}_p (\bm{x}^d - g(\bm{q}))$,
                with a simulation time interval of $\Delta t = $ \SI{1}{s} and $N = 150$ iterations.
                The plots show the six Cartesian dimensions of the error $\bm{x}^d - g(\bm{q})$ for each iteration.
        }
        \label{fig:solver_convergence}
\end{figure*}

In this experiment we analyze our control scheme from
Fig.~\ref{fig:fd_ik_solver} for a distant IK target, using both the Jacobian transpose method and our
conditioned inertia method.
The distant target represents a step in Cartesian space, as illustrated in Fig.~\ref{fig:robot_configurations} for experiment B.
A common use case is low-frequency sample targets.
Since the speed of convergence in both systems
(\ref{eq:wolovich_dynamic_system}) and (\ref{eq:fd_ik_solver}) can be tweaked
with the gains $\bm{K}$ and $\bm{K}_p, \bm{K}_d$ respectively,
we chose the following set of parameters to make the solvers comparable:
\begin{equation}
        \label{eq:convergence_parameters}
        \begin{aligned}
                \Delta t = 1, \bm{K}_p &= \text{diag}(\left[ 1, 1, 1,~0.1, 0.1, 0.1\right]), \\
                \bm{K}_d &= \bm{0},~ \bm{K} = \alpha \bm{I}_6,
        \end{aligned}
\end{equation}
where $\alpha$ is a scaling factor built from the mean matrices from Fig.~\ref{fig:homogenization_statistics} (top row, left + right)
\begin{equation}
        \label{eq:alpha}
                \alpha = \frac{\text{mean(diagonal}(\bm{J} \bm{H}^{-1} \bm{J}^T))}
                {\text{mean(diagonal}( \bm{J} \bm{J}^T))} \approx 0.7885
\end{equation}
to obtain an identical mapping $\bm{f} \rightarrow \ddot{\bm{x}}$ (considering the average of the main diagonal).
Fig.~\ref{fig:solver_convergence} shows the systems' step response to a sudden Cartesian offset.
Both systems approach the goal
state, as indicated by the vanishing errors.
However, there is a big difference in the intermediate solutions:
The Jacobian transpose method overshoots in four out of six dimensions,
and looses track of $\epsilon_{ry}$, which does not flat out for the interval observed.
In comparison, our proposed method converges stronger
and maintains the rotational errors constant throughout the path to the target.

\subsection{Interpolation Quality}
\begin{figure}
        \centering
        \includegraphics[width=0.5\textwidth]{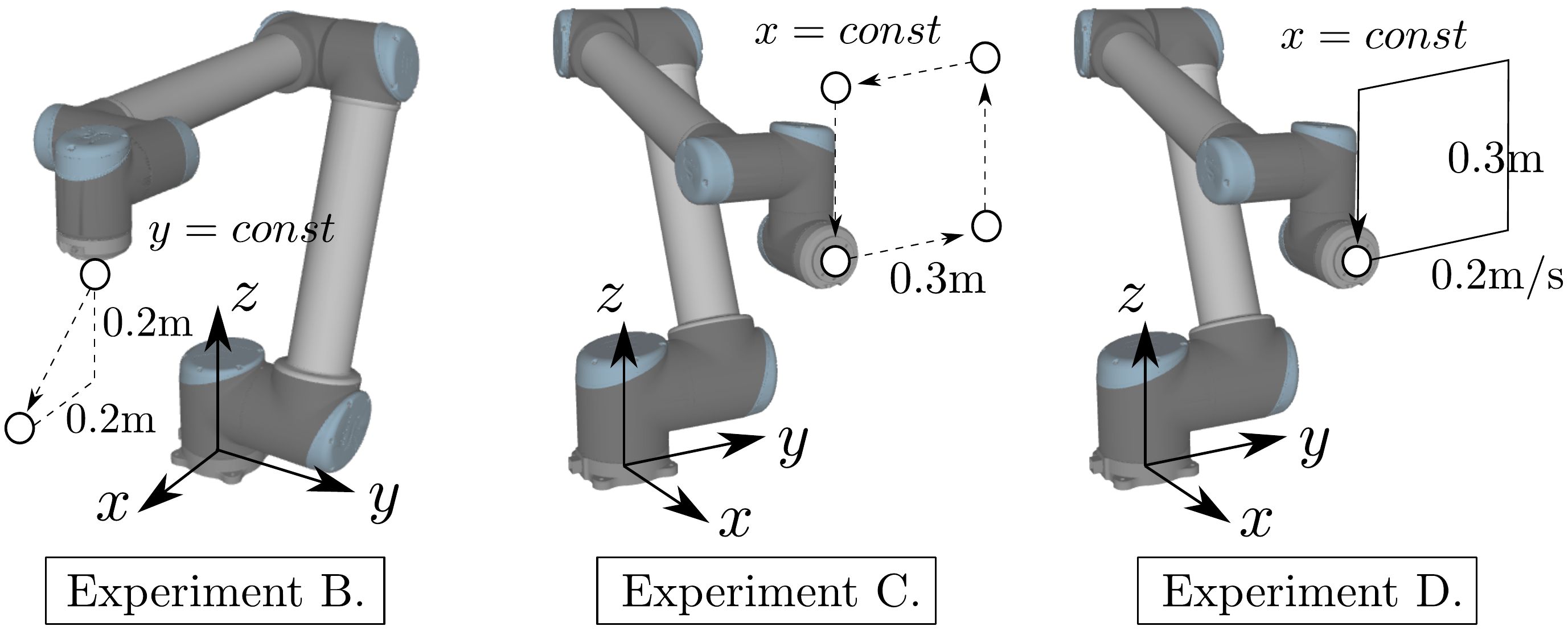}
        \caption{Robot starting configurations and target motion for three experiments.}
        \label{fig:robot_configurations}
\end{figure}

In the next experiment we analyze the interpolation quality of intermediate states.
For this purpose we interpolated to four targets, representing the corners of a square.
The starting configuration is illustrated in Fig.~\ref{fig:robot_configurations} for experiment C.
We chose $\Delta t = \SI{0.1}{s}$ and the number of iterations $N = 50$ for each step.
Starting at all four corners we computed $1000$ steps of intermediate states, for each taking the next
counter clockwise corner as fixed target $\bm{x}^d$.
\begin{figure}
        \centering
        \begin{subfigure}[b]{0.35\textwidth}
                \includegraphics[width=1.0\textwidth]{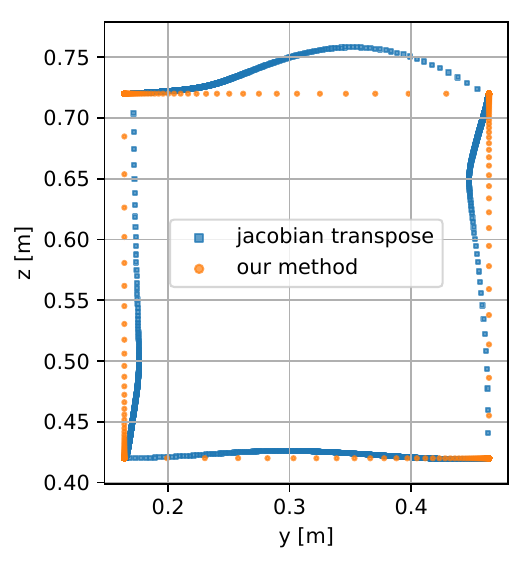}
        \end{subfigure}
        \begin{subfigure}[b]{0.5\textwidth}
                \includegraphics[width=1.0\textwidth]{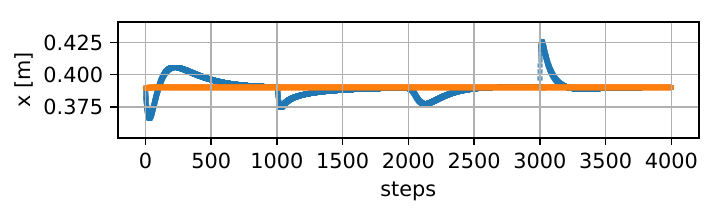}
        \end{subfigure}
        \caption{
                \textbf{Experiment C}. Analysis of path accuracy during interpolation to the corners of a square. We use the
                parameters from (\ref{eq:convergence_parameters}) with $\Delta t = $
                \SI{0.1}{s}. The Cartesian end~effector positions are computed
                with the robots forward kinematics $g(\bm{q})$ for each step.
        }
        \label{fig:solver_interpolation}
\end{figure}
Fig.~\ref{fig:solver_interpolation} shows the results.
The plots demonstrate that our proposed method generates goal-directed, intermediate solutions.
In comparison, the Jacobian transpose method leads to distorted paths in Cartesian space.
Note also that our method converges in less iterations, as indicated with bigger spaces between individual points.

\subsection{Motion tracking}
In this experiment, we apply our method to the tracking of a moving target that
follows the square from Fig.~\ref{fig:robot_configurations} for experiment D
with a constant speed of \SI{0.2}{m/s}.
\begin{figure}
        \centering
        \begin{subfigure}[b]{0.35\textwidth}
                \includegraphics[width=1.0\textwidth]{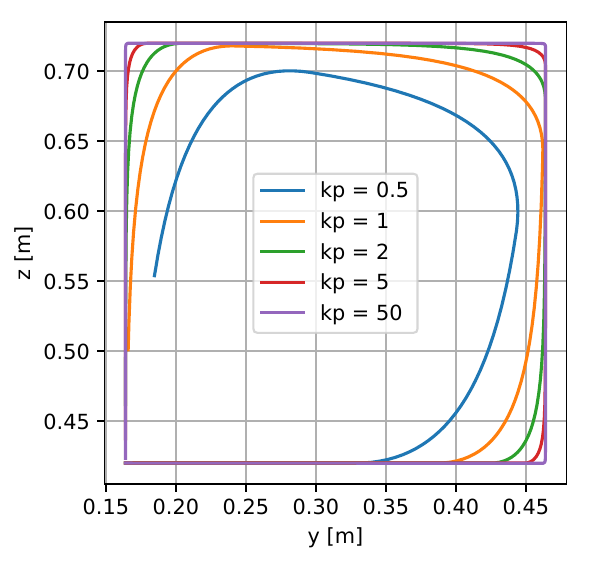}
                \label{fig:motion_tracking_y-z} 
        \end{subfigure}
        \begin{subfigure}[b]{0.5\textwidth}
                \includegraphics[width=1.0\textwidth]{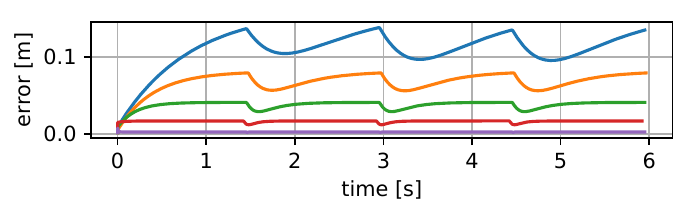}
                \label{fig:motion_tracking_pos_error}
        \end{subfigure}
        \begin{subfigure}[b]{0.5\textwidth}
                \includegraphics[width=1.0\textwidth]{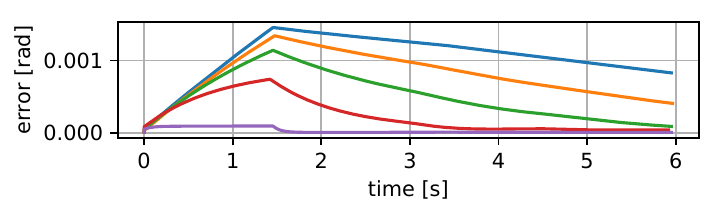}
                \label{fig:motion_tracking_rot_error}
        \end{subfigure}
        \caption{
                \textbf{Experiment D}. Analysis of tracking performance for a
                square-moving target of \SI{0.2}{m/s}. The
                family of curves is computed by varying $k_p$ in the gain
                matrix $\bm{K}_p = k_p \text{diag}(\left[ 1, 1, 1,~0.1, 0.1,
                0.1\right])$. We chose $\Delta t = \SI{0.1}{s}$ and $N = 10$ as IK solver parameters.
                Plot (a) shows the end~effector's absolute motion in Cartesian space. Plot (b)
                shows the according translational error $\lVert \left[ \epsilon_x ~
                \epsilon_y ~ \epsilon_z \right]^T \rVert$ and rotational error $\lVert
                \left[ \epsilon_{rx} ~ \epsilon_{ry} ~ \epsilon_{rz} \right]^T \rVert$
                to the moving target.
        }
        \label{fig:motion_tracking}
\end{figure}
During the execution we sample the moving target with a frequency of \SI{100}{Hz}.
Fig.~\ref{fig:motion_tracking} shows the tracking performance for individual
gains $k_p$, which we use as a multiplicative factor for $\bm{K}_p$ from the previous experiments.
The results show the intended behavior: By increasing the gains we transition from a
smooth but delayed controller to a fast and exact IK solver.
As an example, users may prefer $k_p = 5$ over $k_p = 50$ for smoothed corners for their robot control.

\section{DISCUSSION}
\label{sec:discussion}
\subsection{Virtual Dynamics}
To obtain (\ref{eq:simplified_system}), we neglected non-linear Coriolis and
gravity terms in favor of a simple, unbiased mapping from Cartesian
end~effector offset to joint accelerations.
There are, however, scenarios in which these terms could provide beneficial features.
For redundant manipulators, gravity and Coriolis terms could be used to gain
more control over links and joints in the nullspace, e.g. to implement a
natural behavior of moving the elbow.
For manipulators with six or less degrees of freedom, however, these terms
add computational complexity without real advantages.

Our algorithm does not accumulate velocity during time integration, such that
the system always starts anew in each control cycle with instantaneous
accelerations, having the benefit that no damping is required to avoid overshooting.
Also note, that the dynamic parameters from (\ref{eq:mass_and_inertia}),
which we recommend as default values for any robotic manipulator using our IK
algorithm, are not necessarily realistic, and are, in fact, deliberately
off from the real system's mass and inertia distribution.
In this regard, our used term \textit{twin} refers to the robot kinematics only.
The dynamics are virtually conditioned to obtain an IK solver with the features as described in this work.

\subsection{Controller Gains}
We chose the controller gains empirically for our experiments.
The configuration from Fig.~\ref{fig:motion_tracking} represents a good
starting point for most use cases.
The magnitude of the controller gains directly corresponds to how fine grained
the solver interpolates between the sampled targets.
Users can easily adapt this responsiveness according to their use case through a
single factor ($k_p$ from Fig.~\ref{fig:motion_tracking}).
Additionally, the relative difference between the individual elements in the diagonal gain matrix $\bm{K}_p$ offers the
possibility to adjust how translation is weighed against orientation, and also
which Cartesian dimension gets priority in underactuated system configurations.

\subsection{Efficiency}
Computing the inverse of the manipulator's mass matrix
in each controller cycle introduces additional computational cost.
However, we did not experience this to become a performance issue, even for
solver iteration rates of \SI{10}{kHz}, which we tested on an Intel\textsuperscript{\textregistered} Core\texttrademark~i7-4900MQ.
Experiment~C showed that our solver converges in substantially less iterations.
Weighing this advantage against the more complex computation scheme in terms of
overall convergence time could be subject to further study.

\section{CONCLUSIONS}   %
\addtolength{\textheight}{-12.5cm}   %

In this work we presented a new method for sample-based motion tracking for robotic manipulators.
The core concept of our approach bases on including the mass matrix of a virtually
conditioned twin of the manipulator into the control loop for solving IK.
Deriving our concept from the view of manipulator forward dynamics simulations, we offer
an intuitive solution to achieve task space homogenization despite varying robot joint configurations,
for which we provided an empirical proof on the UR10 robot.
We tested and compared our method in experiments with the UR10 to the well-known Jacobian transpose method.
The results showed that our solution outperforms the Jacobian transpose method
in terms of convergence, path accuracy and interpolation quality.
Users may configure the algorithm to anything from a smooth
controller to a fast and accurate IK solver, also for applications outside the
context of robot motion tracking.

\section*{ACKNOWLEDGMENT}
This work was supported in part by the European Community
Seventh Framework Program under Grant no.~608849~(EuRoC Project).

\printbibliography

\end{document}